\documentclass[11pt]{article}

\usepackage[]{acl}

\usepackage{times}
\usepackage{latexsym}

\usepackage[T1]{fontenc}

\usepackage[utf8]{inputenc}
\usepackage{verbatim}
\usepackage{booktabs}
\usepackage{microtype}

\usepackage{inconsolata}

\usepackage{graphicx}

\usepackage{graphicx}
\usepackage{amsmath}
\usepackage{textgreek}
\usepackage{tikz}
\usepackage{multirow}
\usetikzlibrary{arrows.meta,shapes.geometric,shapes.symbols,shadows}

\usetikzlibrary{shapes, arrows, positioning, calc}
\usepackage{pgfplots}
\pgfplotsset{compat=1.17}
\usepackage{subcaption} 
\usepackage{microtype}
\usepackage[utf8]{inputenc}
\usepackage{newunicodechar}
\newunicodechar{̂}{\^}

\title{R2T: Rule-Encoded Loss Functions for Sequence Tagging in Low-Resource Languages}

\author{
 \textbf{Mamadou K. KEITA\textsuperscript{1}},
 \textbf{Christopher Homan\textsuperscript{1}},
 \textbf{Sebastien Diarra\textsuperscript{2}} 
\\
 \textsuperscript{1}Rochester Institute of Technology,
 \textsuperscript{2}RobotsMali
}

\begin{document}
\maketitle
\begin{abstract}
We introduce Rule-to-Tag (R2T), a framework that turns linguistic rules into the training signal for neural sequence taggers in low-resource languages. R2T encodes lexical, morphological, and syntactic rules as differentiable loss terms, so that a tagger learns from rules and unlabeled text, with no labeled training data. It also adds an out-of-vocabulary (OOV) loss term that discourages confident predictions on words that no rule covers. R2T is a first instance of a broader paradigm we call principled learning (PrL): using explicit principles as a learning signal, rather than relying on example-based supervision alone. We evaluate R2T on part-of-speech (POS) tagging for Zarma (Songhay), Bambara (Mande), and French (Romance), and on named entity recognition (NER) for Zarma. On Zarma POS tagging, R2T-BiLSTM uses no labeled training data, yet it reaches 0.968 Macro F1. This score comes within 0.007 of a supervised BiLSTM-CRF trained on 300 labeled sentences (0.975) and exceeds AfriBERTa fine-tuned on the same sentences (0.941). On NER, R2T works well as pre-training: after R2T pre-training, a model fine-tuned on 50 labeled sentences outperforms AfriBERTa fine-tuned on 300 (0.83 vs.\ 0.79 span F1). On Bambara, R2T reaches 0.91 Macro F1 with rules written in about 2.75 hours, while a supervised Masakhane tagger reaches 0.78. We release ZarmaPOS-Bench, a silver-standard Zarma POS corpus, together with ZarmaNER-600 and all trained models, to support future work on under-resourced languages.
\end{abstract}

\section{Introduction}
\label{sec:introduction}

Part-of-speech (POS) tagging supports many NLP applications, from machine translation to syntactic parsing and information extraction. Deep learning models achieve near-perfect accuracy on this task for high-resource languages, but low-resource languages lack the large annotated datasets these models require. For many African languages, even basic sequence tagging tools remain unavailable~\citep{adelani-etal-2022-masakhaner}.

Of the common strategies used to close this gap, transfer learning and cross-lingual projection require parallel data and careful alignment~\citep{das-petrov-2011-unsupervised}, while multilingual transformers depend on large-scale pre-training pipelines, well-matched tokenizers, and compute resources that many communities cannot access~\citep{DBLP:journals/corr/abs-1911-02116}. Purely rule-based taggers handle easy cases but break on ambiguity. Large language models are also impractical for our target setting, as they demand significant compute and hallucinate on languages with little training data.

We propose the \textbf{Rule-to-Tag (R2T)} framework to address these challenges. R2T builds explicit linguistic rules directly into a neural network's training objective through a multi-tiered system of lexical, morphological, and syntactic constraints. This design gives the model a linguistic scaffold that guides learning even when no labeled data exists. R2T also adds an out-of-vocabulary (OOV) loss term, which penalizes confident predictions on words that no rule covers. Such words are common in low-resource settings, where code-switching and borrowed words are frequent.

The core idea behind R2T is simple: experts write down what they already know about a language, and R2T turns that knowledge into the training signal. This idea belongs to a broader paradigm that we call \textbf{principled learning (PrL)}: using explicit principles as a learning signal, rather than relying on example-based supervision alone. R2T is a first, loss-based instance of PrL. Its rules serve as the \emph{primary training signal}, not as a regularization term or a post-processing step. Unlike constraint-based methods that operate outside the model's gradient update~\citep{JMLR:v11:ganchev10a, 10.1007/s10994-012-5296-5}, R2T turns rules into differentiable loss terms, so it can train a tagger without any labeled training data.

We evaluate R2T mainly on Zarma, a Songhay language that had no large-scale POS corpus before this work. We also test R2T on Bambara (Mande family), where it reaches 0.91 Macro F1 with rules written in about 2.75 hours and outperforms a supervised tagger (0.78). Finally, we stress-test R2T on French under simulated low-resource conditions to see whether it transfers beyond African languages. Our work addresses two research questions:

\begin{description}
    \item[RQ1:] Can a model trained with linguistic rules and unlabeled text outperform large pre-trained models fine-tuned on a small labeled dataset?
    \item[RQ2:] How effectively does R2T pre-training reduce the labeled data needed for complex tasks like NER?
\end{description}

\noindent Our contributions are:

\begin{enumerate}
    \item \textbf{The R2T framework, a first instance of PrL:} We introduce a training framework that turns a multi-tiered linguistic rule system into differentiable loss terms, including an OOV term that penalizes confident predictions on tokens that no rule covers.

    \item \textbf{POS tagging without labeled training data:} R2T-BiLSTM reaches 0.968 Macro F1 (98.2\% accuracy) without labeled training data. It comes within 0.007 of a supervised BiLSTM-CRF trained on 300 labeled sentences (0.975) and outperforms AfriBERTa fine-tuned on the same sentences (0.941).

    \item \textbf{Data-efficient pre-training for NER:} After R2T pre-training, a model fine-tuned on only 50 labeled NER sentences outperforms AfriBERTa fine-tuned on 300 (0.83 vs.\ 0.79 span F1).

    \item \textbf{Cross-lingual results:} R2T transfers to Bambara (0.91 vs.\ 0.78 Macro F1 against a supervised Masakhane tagger) and to French under simulated data scarcity (0.941 Macro F1 with 1{,}000 unlabeled sentences).

    \item \textbf{New resources:} We release ZarmaPOS-Bench, a silver-standard corpus of about 55{,}000 Zarma sentences; ZarmaNER-600, with 600 gold-standard NER sentences; the gold POS data described in Section~\ref{ssec:data}; pre-trained Zarma FastText embeddings; and all trained models.\footnote{\url{https://huggingface.co/27Group}}
\end{enumerate}

\section{Related Work}
\label{sec:related}

\paragraph{Low-resource POS tagging}
The main obstacle for POS tagging in low-resource languages is the lack of annotated data. Cross-lingual projection transfers supervision from high-resource languages through parallel data or word alignments~\citep{das-petrov-2011-unsupervised, tackstrom-etal-2013-token}. Classic probabilistic taggers such as HMMs or TnT~\citep{brants2000tntstatisticalpartofspeech} can work well, but they lack the contextual power of neural models. \citet{garrette-baldridge-2013-learning} show that small, targeted amounts of annotation, combined with morphological information and type-level constraints, can already produce useful taggers. For African languages, MasakhaPOS provides POS data for 20 languages, including Bambara~\citep{dione-etal-2023-masakhapos}. Multilingual models such as XLM-RoBERTa~\citep{DBLP:journals/corr/abs-1911-02116} and African-centric models such as AfriBERTa~\citep{ogueji-etal-2021-small} are strong baselines, but their tokenizers can fit low-resource languages poorly~\citep{rust-etal-2021-good}.

\paragraph{Learning from constraints}
A long line of work adds prior knowledge to machine learning models. Posterior regularization~\citep{JMLR:v11:ganchev10a} and generalized expectation criteria~\citep{JMLR:v11:mann10a} use constraints to guide model posteriors, often through an auxiliary optimization step, and generalized expectation can even train a classifier from labeled features without any labeled instances~\citep{druck2008learning}. Constrained conditional models enforce rules during inference~\citep{10.1007/s10994-012-5296-5}, while data programming turns noisy, heuristic labeling functions into training labels~\citep{ratner2017dataprogrammingcreatinglarge}. For neural networks, differentiable losses can encode logical constraints~\citep{xu2018semantic}, and constraint losses can train networks without labels in some settings~\citep{stewart2017label}. R2T builds on several of these ideas. Its ambiguous-tag loss follows the partial-label idea behind the type constraints of \citet{tackstrom-etal-2013-token}; its OOV term is a confidence penalty~\citep{pereyra2017regularizingneuralnetworkspenalizing} that R2T applies only to tokens that no rule covers; and its syntactic loss replaces the usual CRF layer~\citep{huang2015bidirectionallstmcrfmodelssequence} with a soft, differentiable alternative.

\paragraph{Positioning R2T and PrL}
Most of this work uses knowledge to regularize or constrain models that learn from labels, or it turns rules into noisy labels. A few works already learn from constraints alone in specific settings~\citep{stewart2017label, druck2008learning}. PrL, which we define in Section~\ref{sec:introduction}, names this idea and separates it from the channel that delivers the principles, whether a loss term, a reward, or another signal. PrL is also narrower than informed machine learning~\citep{vonrueden2023informed}, which surveys the many ways to add prior knowledge to learning systems: in PrL, the principles are the main learning signal, not an addition to labeled data. Within this view, R2T contributes a tiered, rules-only training objective for neural sequence taggers, results in truly low-resource languages, and new resources for Zarma.

\section{The R2T Approach}

R2T is a hybrid framework that combines the contextual learning of neural networks with a multi-tiered system of linguistic knowledge. Instead of applying rules as a post-processing step, R2T builds them into the model's training objective through a rule-informed loss function. This loss pushes the model to follow known linguistic facts, and it discourages confident predictions on words that no rule covers. Figure~\ref{fig:r2t_new} gives an overview.

R2T has three components. First, a neural tagger captures contextual patterns in text. Second, a multi-tiered rule system provides explicit linguistic knowledge. Finally, a rule-informed loss function connects the two and guides the tagger toward grammatical predictions. The following subsections describe each component.

\begin{figure*}[t]
\tiny
\centering
\begin{tikzpicture}[
    node distance=1.2cm and 1.5cm,
    >=stealth,
    block/.style={rectangle, rounded corners, draw=black, fill=blue!10, minimum height=1cm, align=center},
    rule/.style={rectangle, draw=black, fill=gray!15, minimum height=0.8cm, align=center},
    loss/.style={rectangle, draw=black, fill=red!10, minimum height=0.8cm, align=center},
    arrow/.style={->, thick}
]

\node[rule] (tier1) {Tier 1: Unambiguous Lexicon};
\node[rule] (tier2) [below=0.5cm of tier1] {Tier 2: Ambiguous Lexicon};
\node[rule] (tier3) [below=0.5cm of tier2] {Tier 3: Morphological Rules};
\node[rule] (tier4) [below=0.5cm of tier3] {Tier 4: Syntactic Rules};

\node[loss] (l_lex) [right=2.5cm of tier2] {$\mathcal{L}_{\text{lex}}$ \\ (Lexical Loss)};
\node[loss] (l_syn) [right=2.5cm of tier4] {$\mathcal{L}_{\text{syn}}$ \\ (Syntactic Loss)};

\node[block] (model) [right=2.5cm of l_lex] {Neural Tagger \\ (BiLSTM or Transformer) \\ $p_\theta(y_i \mid x_{1:N})$};

\node[loss] (l_dist) [above=0.8cm of model] {$\mathcal{L}_{\text{dist}}$ \\ (Distributional)};
\node[loss] (l_oov) [below=0.8cm of model] {$\mathcal{L}_{\text{oov}}$ \\ (OOV)};

\node[block, fill=green!15] (output) [right=2.5cm of model] {Predicted Tags \\ $\hat{y}_{1:N}$};

\draw[arrow] (tier1.east) -- (l_lex.west);
\draw[arrow] (tier2.east) -- (l_lex.west);
\draw[arrow] (tier3.east) -- (l_lex.west);
\draw[arrow] (tier4.east) -- (l_syn.west);

\draw[arrow, line width=1.5pt, red!60!black] (l_lex.east) -- (model.west) node[midway, above, sloped, black] {Guidance};
\draw[arrow, line width=1.5pt, red!60!black] (l_syn.east) -- (model.west);
\draw[arrow, line width=1.5pt, red!60!black] (l_dist.south) -- (model.north);
\draw[arrow, line width=1.5pt, red!60!black] (l_oov.north) -- (model.south);

\draw[arrow, line width=1.5pt, blue!60!black] (model.east) -- (output.west) node[midway, above, black] {Prediction};

\end{tikzpicture}
\caption{The R2T framework. R2T turns a multi-tiered rule system into loss terms that guide the training of a neural sequence tagger. The lexical and syntactic losses enforce known grammar, while the distributional and OOV losses regularize the tagger's predictions.}
\label{fig:r2t_new}
\end{figure*}

\subsection{Neural Architecture}

R2T uses a standard neural architecture for sequence tagging. For each token in a sentence, we build a representation from two sources. First, we use pre-trained word embeddings, for example from FastText~\citep{bojanowski-etal-2017-enriching} or any other embedding model. These embeddings carry distributional information, which matters in low-resource settings, where a model cannot learn such representations from a small annotated dataset alone. Second, to handle morphological variation and unseen words, we build a character-level representation for each token.

We feed the characters of each token into a character-level bidirectional long short-term memory network (BiLSTM) and concatenate its final forward and backward hidden states. This lets the model build representations for unseen words from their sub-word structure, a method that works well in many tagging tasks~\citep{lample-etal-2016-neural}.

We then concatenate the word embedding and the character-level embedding. This combined vector is the input to the main token-level encoder, which is either a BiLSTM or a Transformer. The encoder reads the whole sentence and produces a context-aware representation for each token. Finally, a linear layer and a softmax turn this representation into a probability distribution over the tagset. Figure~\ref{fig:bilstm_arch} in Appendix~\ref{sec:technical_details} shows the BiLSTM version.

\subsection{A Multi-Tiered Linguistic Rule System}

The key design choice in R2T is to organize rules into tiers that give the neural model a scaffold for learning. The tiers run from high-confidence facts to general heuristics, which allows graded guidance. We define four tiers and illustrate each with a Zarma example.

\paragraph{Tier 1: Unambiguous lexical rules} This tier is the base of the rule system. It lists words that always take a single POS tag, typically high-frequency function words (such as pronouns and determiners) and core vocabulary whose tag does not change across contexts. For example, Tier~1 lists the Zarma verb \textit{wani} (``to play'') as \textsc{verb}.

\paragraph{Tier 2: Ambiguous lexical rules} Lexical ambiguity is a key challenge in many languages, especially low-resource ones. This tier lists words that can take several POS tags. For example, Zarma \textit{no} can be an auxiliary or the main verb ``to give'', so Tier~2 lists it as \textsc{aux} or \textsc{verb}. Instead of forcing a single tag, the tier gives the model a small set of valid options, and the neural tagger uses context to choose among them.

\paragraph{Tier 3: Morphological rules} To help with unseen words, this tier captures common morphological patterns. These rules usually rely on suffixes or prefixes and suggest a likely tag. For example, the Zarma definite suffixes \textit{-a} and \textit{-o} suggest that a word is a noun. The tier provides a fallback when no lexical entry exists for a word.

\paragraph{Tier 4: Syntactic rules} This tier models local grammar by scoring transitions between adjacent POS tags. We store these rules as a matrix of transition validities; for example, the Zarma rules mark a pronoun followed by an auxiliary as a valid transition. The tier helps the model produce coherent, grammatical tag sequences.

\subsection{Rule-Informed Loss Function}
\label{sec:adaptive_loss}

A multi-part loss function connects the tagger and the rules. It turns the rule system into training objectives that guide what the model learns. In our unsupervised setting, these rule-based losses are the \emph{primary} training signal, not regularization terms added to a supervised objective. This is what allows R2T to train without labeled training data.

The total loss $\mathcal{L}_{\text{R2T}}$ is a weighted sum of four components:
\begin{equation}
    \mathcal{L}_{\text{R2T}} = \alpha\mathcal{L}_{\text{lex}} + \beta\mathcal{L}_{\text{syn}} + \gamma\mathcal{L}_{\text{dist}} + \delta\mathcal{L}_{\text{oov}}
\end{equation}
where $\alpha$, $\beta$, $\gamma$, and $\delta$ are hyperparameters that balance the contribution of each term. We set them on the Rule-Dev split (Section~\ref{ssec:data}).

\paragraph{Lexical loss ($\mathcal{L}_{\text{lex}}$)} This term enforces the lexical and morphological rules (Tiers 1--3). For a token $x_i$ to which the rules assign a single tag $y_i$, the loss is the standard negative log-likelihood:
\begin{equation}
    \mathcal{L}_{\text{lex-unambig}} = -\log p_\theta(y_i \mid x_{1:N})
\end{equation}
where $p_\theta(y_i \mid x_{1:N})$ is the probability that the tagger assigns to tag $y_i$ at position $i$, given the sentence $x_{1:N}$.

For a token with a set of valid tags $Y_{\text{ambig}}$, we sum the probabilities of all valid tags instead. This pushes probability mass into the valid set without forcing an early choice among its members:
\begin{equation}
    \mathcal{L}_{\text{lex-ambig}} = -\log\left(\sum_{y' \in Y_{\text{ambig}}} p_\theta(y' \mid x_{1:N})\right)
\end{equation}

\paragraph{Syntactic loss ($\mathcal{L}_{\text{syn}}$)} This term enforces the Tier~4 syntactic rules. We define a transition invalidity matrix $M$, where $M_{jk} = 1 - \text{validity}(tag_j \rightarrow tag_k)$. We compute the loss for a sentence as the average penalty over adjacent pairs of predictions:
\begin{equation}
    \mathcal{L}_{\text{syn}} = \frac{1}{N-1} \sum_{i=1}^{N-1} \mathbf{p}_i^T M \mathbf{p}_{i+1}
\end{equation}
where $\mathbf{p}_i$ is the vector of tag probabilities at position $i$. This term discourages grammatically invalid tag sequences.

\paragraph{Distributional loss ($\mathcal{L}_{\text{dist}}$)} This term is the Kullback-Leibler (KL) divergence~\citep{shlens2014noteskullbackleiblerdivergencelikelihood} between the model's average predicted tag distribution and a uniform distribution. It keeps the model from collapsing onto a few frequent tags. Real tag frequencies are very skewed, so we use this term only as a weak guard against collapse, with a small weight ($\gamma = 0.02$), and not as a target distribution.

\paragraph{OOV loss ($\mathcal{L}_{\text{oov}}$)} In R2T, an out-of-vocabulary (OOV) token is a token that no Tier~1--3 rule covers. This differs from the usual meaning of a word unseen in training. For such a token $x_{\text{oov}}$, we want the model to avoid a confident and likely incorrect prediction. We therefore penalize the model when its output distribution $\mathbf{p}_{\text{oov}}$ moves far from a uniform distribution $\mathcal{U}$, which we measure with the KL divergence:
\begin{equation}
    \mathcal{L}_{\text{oov}} = D_{\text{KL}}(\mathbf{p}_{\text{oov}} || \mathcal{U}) = \sum_{j=1}^{|T|} p_j \log\left(\frac{p_j}{1/|T|}\right)
\end{equation}
where $|T|$ is the number of tags in the tagset. This term is a confidence penalty that applies only where the rules give no guidance. It pushes the model toward flatter distributions on uncovered tokens, while the lexical loss keeps its predictions on covered tokens confident.

\paragraph{Differences from prior work} Together, these components make R2T an end-to-end differentiable system, in which rules are not heuristics or constraints applied after the fact but part of the training objective itself. This design distinguishes R2T from earlier constraint-based approaches that operate outside the model's gradient update~\citep{JMLR:v11:ganchev10a, 10.1007/s10994-012-5296-5}. In those methods, rules guide inference or regularize posteriors through auxiliary optimization. In R2T, the rules \emph{are} the supervision: they replace labeled training data entirely in our unsupervised setup.

\section{Experiments}
\label{sec:experiments}

We evaluate R2T on POS tagging for Zarma, Bambara, and French, and on NER for Zarma. Our goals are threefold: (1)~test whether R2T, trained only with linguistic rules and unlabeled text, can outperform pre-trained models fine-tuned on small annotated datasets; (2)~analyze how the neural architecture interacts with our rule-based objective; and (3)~test whether R2T transfers to other languages.

\subsection{Data}
\label{ssec:data}

Our main experiments target Zarma, a Songhay language spoken in Niger with very little publicly available annotated data.

\paragraph{POS tagging} For unsupervised training, we use 25,000 sentences from the Zarma grammatical error correction (GEC) dataset~\citep{keita2025grammaticalerrorcorrectionlowresource}, and we train FastText embeddings on the full corpus. Three experts annotated the gold data (Krippendorff's $\alpha = 0.93$ for POS and $0.97$ for NER). We define four disjoint splits: (i)~\textbf{Unlabeled} (25k sentences) for unsupervised R2T training; (ii)~\textbf{Rule-Dev} (100 gold sentences) for rule refinement and hyperparameter tuning only; (iii)~\textbf{Gold-Train} (300 gold sentences) for supervised baselines and supervised fine-tuning (SFT); and (iv)~\textbf{Gold-Test} (1,000 gold sentences), held out until the final evaluation. No sentence appears in more than one split, and we froze all rules and hyperparameters on Rule-Dev before evaluating on Gold-Test. R2T therefore uses no labeled data for training, but it does use the 100 Rule-Dev sentences to refine its rules. We release these splits with ZarmaPOS-Bench, which we build from Feriji~\citep{keita-etal-2024-feriji} and describe in Appendix~\ref{sec:zarmapos_bench}.

\paragraph{NER} We created ZarmaNER-600, a new gold-standard dataset of 600 manually annotated sentences with four entity types: persons (\textsc{per}), locations (\textsc{loc}), organizations (\textsc{org}), and dates (\textsc{date}), in the standard BIO scheme. We use the first 300 sentences to train supervised baselines, the next 100 as a held-out test set, and the last 50 sentences of the training split for SFT.

\paragraph{Tagset} For Zarma, we use eight POS tags: \textsc{pron}, \textsc{noun}, \textsc{verb}, \textsc{adj}, \textsc{aux}, \textsc{part}, \textsc{det}, and \textsc{punct}.

\subsection{Rules}
\label{ssec:rules}

We built a multi-tiered rule system for Zarma with 20 grammar rules, drawn from existing linguistic documentation and feedback from native speakers. The rules follow three design guidelines: (1)~prioritize high-frequency, low-ambiguity words; (2)~list ambiguous words explicitly; and (3)~refine rules step by step based on model errors on Rule-Dev. We compiled a Tier~1 lexicon of unambiguous words, a Tier~2 lexicon of ambiguous words, morphological patterns such as the definite suffixes \textit{-a} and \textit{-o}, and syntactic constraints such as a pronoun followed by an auxiliary. A trained native speaker and an NLP researcher wrote the rules in about 4 hours. The rules are available in machine-readable JSON format.\footnote{\url{https://huggingface.co/datasets/27Group/ZarmaLanguageRules}}

For NER, we wrote a separate rule set that encodes capitalization heuristics, common title words (such as ``Monsieur'' and ``Madame''), location suffixes, and syntactic contexts that signal entity boundaries. The NER rules use the same four tiers but target entity spans rather than single-token tags.

\subsection{Experimental Setup}
\label{ssec:setup}

We compare seven model configurations for POS tagging and four for NER. All baselines use the same \texttt{wordpunct} tokenization to eliminate tokenizer mismatches.

\paragraph{BiLSTM-CRF} A supervised baseline with a CRF layer for structured prediction, trained from scratch on the full 300-sentence Gold-Train set. Unlike R2T, it uses learned word embeddings rather than FastText. Appendix~\ref{ssec:appendix_arch} gives architecture details.

\paragraph{R2T-BiLSTM} Our main model, trained for 30 epochs on 25,000 unlabeled sentences with the rule-informed loss (Section~\ref{sec:adaptive_loss}).

\paragraph{R2T-Transformer} An architectural variant that replaces the BiLSTM encoder with a Transformer encoder (10 layers, 6 heads, 768 hidden units, 3072 feed-forward units) but uses the same rules and loss function. This configuration matches the size of AfriBERTa-large~\citep{ogueji-etal-2021-small}.

\paragraph{R2T-Transformer SFT-50} R2T-Transformer after 20 more epochs of supervised fine-tuning on 50 Gold-Train sentences with a standard cross-entropy loss.

\paragraph{AfriBERTa} An African-centric pre-trained model~\citep{ogueji-etal-2021-small}, fine-tuned on the full 300-sentence Gold-Train set for 10 epochs.

\paragraph{XLM-RoBERTa} A multilingual baseline~\citep{DBLP:journals/corr/abs-1911-02116}, fine-tuned on the full 300-sentence Gold-Train set for 10 epochs.

\paragraph{Rule-based Classifier} Uses only Tiers~1--3 as a per-word lookup: it assigns the unambiguous tag (Tier~1), the most frequent ambiguous tag (Tier~2), or the morphologically suggested tag (Tier~3).

We report Macro F1 (the unweighted average of per-tag F1 scores) as the main metric and word-level accuracy (Acc.) for completeness. All results are averages over 5 random seeds. Appendix~\ref{ssec:appendix_hyper} lists the hyperparameters for all models.

\subsection{POS Tagging Results}
\label{ssec:pos_results}

\begin{table*}[!ht]
\centering
\tiny
\begin{tabular}{l|cccccccc|c|c}
\hline
\textbf{Model} & \textbf{PRON} & \textbf{NOUN} & \textbf{VERB} & \textbf{ADJ} & \textbf{AUX} & \textbf{PART} & \textbf{DET} & \textbf{PUNCT} & \textbf{Macro F1} & \textbf{Acc. (\%)} \\
\hline
BiLSTM-CRF & 0.99\textpm.01 & 0.98\textpm.01 & 0.97\textpm.01 & 0.96\textpm.02 & 0.99\textpm.01 & 0.96\textpm.02 & 0.95\textpm.03 & 1.00\textpm.00 & \textbf{0.975\textpm.01} & 98.8\textpm.1 \\
R2T-BiLSTM & 0.99\textpm.01 & 0.97\textpm.01 & 0.96\textpm.02 & 0.94\textpm.03 & 0.98\textpm.01 & 0.95\textpm.02 & 0.94\textpm.03 & 1.00\textpm.00 & \textbf{0.968\textpm.01} & 98.2\textpm.2 \\
\hline
AfriBERTa (SFT-300) & 0.98\textpm.02 & 0.95\textpm.02 & 0.94\textpm.03 & 0.88\textpm.04 & 0.97\textpm.01 & 0.92\textpm.03 & 0.89\textpm.05 & 1.00\textpm.00 & 0.941\textpm.02 & 96.8\textpm.3 \\
R2T-Transformer SFT-50 & 0.98\textpm.02 & 0.94\textpm.02 & 0.93\textpm.03 & 0.89\textpm.04 & 0.96\textpm.02 & 0.91\textpm.03 & 0.90\textpm.04 & 1.00\textpm.00 & 0.935\textpm.02 & 96.3\textpm.4 \\
R2T-Transformer & 0.96\textpm.03 & 0.87\textpm.04 & 0.86\textpm.04 & 0.74\textpm.06 & 0.92\textpm.03 & 0.84\textpm.05 & 0.80\textpm.06 & 0.98\textpm.01 & 0.852\textpm.04 & 89.8\textpm.8 \\
\hline
Rule-based Classifier & 1.00\textpm.00 & 0.41\textpm.02 & 0.31\textpm.02 & 0.00\textpm.00 & 0.95\textpm.01 & 1.00\textpm.00 & 0.00\textpm.01 & 1.00\textpm.00 & 0.584\textpm.01 & 62.4\textpm.2 \\
XLM-RoBERTa (SFT-300) & 0.40\textpm.08 & 0.45\textpm.07 & 0.38\textpm.09 & 0.27\textpm.11 & 0.41\textpm.08 & 0.39\textpm.08 & 0.30\textpm.12 & 0.70\textpm.05 & 0.413\textpm.08 & 49.1\textpm2.1 \\
\hline
\end{tabular}
\caption{Zarma POS tagging results on the 1,000-sentence Gold-Test set, averaged over 5 seeds. R2T-BiLSTM and R2T-Transformer use no labeled training data. BiLSTM-CRF, AfriBERTa, and XLM-RoBERTa train on the 300 Gold-Train sentences, and R2T-Transformer SFT-50 fine-tunes on 50 of them. Bold marks the best Macro F1 overall and the best without labeled training data.}
\label{tab:results_f1}
\end{table*}

Table~\ref{tab:results_f1} shows the Zarma POS tagging results. Without any labeled training data, R2T-BiLSTM reaches 0.968 Macro F1. This score comes within 0.007 of the supervised BiLSTM-CRF (0.975), which trains on all 300 Gold-Train sentences, and it exceeds AfriBERTa fine-tuned on the same sentences (0.941). In contrast, R2T-Transformer is much weaker in the unsupervised setting (0.852). However, it recovers to 0.935 after fine-tuning on only 50 sentences. XLM-RoBERTa scores far lower (0.413), even on punctuation (0.70 F1). A poor fit between its subword tokenizer and Zarma text may explain part of this gap, but we treat this result with caution.

In answer to RQ1, a model trained only with rules and unlabeled text can outperform large pre-trained models fine-tuned on 300 labeled sentences, and it comes close to a supervised tagger trained on the same data.

\subsection{NER Results}
\label{ssec:ner_results}

\begin{table}[h]
\centering
\small
\setlength{\tabcolsep}{4pt}
\begin{tabular}{l|c|c}
\hline
\textbf{Model} & \textbf{Span F1} & \textbf{Acc. (\%)} \\
\hline
R2T-Transformer SFT-50 & \textbf{0.83\textpm.02} & 89.9\textpm.5 \\
AfriBERTa (SFT-300) & 0.79\textpm.03 & 88.9\textpm.6 \\
R2T-BiLSTM & 0.61\textpm.04 & 75.4\textpm.9 \\
R2T-Transformer & 0.53\textpm.05 & 67.4\textpm1.2 \\
\hline
\end{tabular}
\caption{Zarma NER results on the 100-sentence test set (span-level F1), averaged over 5 seeds.}
\label{tab:ner_results_f1}
\end{table}

Table~\ref{tab:ner_results_f1} reports span-level F1 for Zarma NER. A span counts as correct only if both its entity type and its boundaries are correct, so span F1 is stricter than token-level accuracy. We focus on span F1, since \textsc{O} tags dominate token-level accuracy.

Without labeled data, the R2T models reach only modest scores (0.53--0.61), because rules are hard to apply to a sparse task in which most tokens are \textsc{O}. However, R2T works well as pre-training. R2T-Transformer SFT-50, which we pre-train with rules and then fine-tune on 50 gold sentences, reaches 0.83 span F1 and outperforms AfriBERTa fine-tuned on 300 sentences (0.79). The test set has only 100 sentences, so we read this gap as a sign of data efficiency rather than a precise measure of it.

In answer to RQ2, R2T pre-training reduces the labeled data needed for NER in our setting: with it, 50 labeled sentences are enough to outperform AfriBERTa trained on 300.

\subsection{Cross-Lingual Generalization}
\label{ssec:cross_lingual}

To test whether R2T transfers to other languages, we evaluate it on two more languages from different families: Bambara (Mande) and French (Romance, under simulated low-resource conditions). For Bambara, we pair a pre-trained T5-small encoder~\citep{2020t5} with our BiLSTM tagger head, and we compare against the released Masakhane AfroXLMR tagger, a supervised model fine-tuned on annotated Bambara data. We evaluate both on 1,000 Bambara sentences. For French, we report R2T-BiLSTM only, since the other baselines are already heavily pre-trained on French. Appendices~\ref{sec:bambara_experiment} and~\ref{ssec:french_experiment} give full details.

\begin{table}[h]
\centering
\small
\begin{tabular}{lp{2.0cm}cc}
\hline
\textbf{Language} & \textbf{Model} & \textbf{Macro F1} & \textbf{Acc. (\%)} \\
\hline
\multirow{2}{*}{Bambara} & R2T-BiLSTM + T5 & \textbf{0.91\textpm.02} & 92.7\textpm.4 \\
 & AfroXLMR & 0.78\textpm.03 & 82.5\textpm.7 \\
\hline
\multirow{1}{*}{French} & R2T-BiLSTM & \textbf{0.941\textpm.02} & 95.8\textpm.3 \\
\hline
\end{tabular}
\caption{Cross-lingual POS tagging results, averaged over 5 seeds. For Bambara, R2T uses 864 unlabeled sentences and rules written in about 2.75 hours. For French, it uses 1,000 unlabeled sentences and 15 rules written in under one hour.}
\label{tab:cross_lingual}
\end{table}

Table~\ref{tab:cross_lingual} summarizes the results. On Bambara, R2T outperforms the supervised Masakhane tagger by 0.13 Macro F1 (0.91 vs.\ 0.78), using rules written in about 2.75 hours from Daba morphemic patterns~\citep{daba}. On French, R2T reaches 0.941 Macro F1 with only 1,000 unlabeled sentences and 15 basic rules. Together, these results suggest that R2T transfers to other languages when basic linguistic rules are available.

\section{Analysis and Discussion}
\label{sec:analysis}

\begin{table*}[ht]
\centering
\scriptsize
\begin{tabular}{p{2.2cm}|p{2.5cm}|p{4cm}|p{5.5cm}}
\hline
\textbf{Model} & \textbf{Error Type} & \textbf{Example} & \textbf{Analysis} \\
\hline \hline

XLM-RoBERTa & Tokenization mismatch & \textit{Ni neera moo.} \newline Tokens: [\textit{Ni, neera, moo.}] \newline Tags: [\textsc{pron, verb, verb}] & The tokenizer fails to separate punctuation from the preceding word, so ``moo.''\ becomes a single token. This guarantees errors on every sentence and confuses the model on the word itself. \\
\hline

AfriBERTa & Lexical ambiguity & \textit{Ay no a se moo.} \newline Pred: \textit{no} $\rightarrow$ \textsc{aux} \newline Gold: \textit{no} $\rightarrow$ \textsc{verb} & The model defaults to the more frequent auxiliary sense of ``no'' and fails to use the syntactic context (Subject \_ Object) to identify it as the main verb ``to give.'' R2T's Tier~2 rules explicitly encode this ambiguity. \\
\hline

BiLSTM-CRF & Unseen word & \textit{...care fassaro te.} \newline Pred: \textit{fassaro} $\rightarrow$ \textsc{noun} \newline Gold: \textit{fassaro} $\rightarrow$ \textsc{verb} & The word ``fassaro'' (to explain) never appears in the 300 training sentences, so the model guesses from context and morphology. For tokens that no rule covers, R2T's OOV loss instead discourages confident guesses. \\
\hline

R2T-BiLSTM & Adjective--noun confusion & \textit{Iri ya boro yaaje no.} \newline Pred: \textit{yaaje} $\rightarrow$ \textsc{noun} \newline Gold: \textit{yaaje} $\rightarrow$ \textsc{adj} & A rare error on a complex adjective in the pattern \textsc{pron aux pron adj aux}. The rules handle most adjective cases, but this specific construction proved difficult without explicit labeled examples. \\
\hline

\end{tabular}
\caption{Representative errors across model families. Each row shows a different error type; Appendix~\ref{sec:additional_figures} gives the full error analysis.}
\label{tab:error_analysis}
\end{table*}

\paragraph{Rules as an alternative to annotation}
R2T-BiLSTM outperforms AfriBERTa fine-tuned on 300 expert-annotated sentences (Macro F1: 0.968 vs.\ 0.941) without any labeled training data. The human effort behind the two approaches is also similar. Writing the Zarma rules took a native speaker and an NLP researcher about 4 hours, and refining them used the 100 Rule-Dev sentences. By comparison, annotating the 300 Gold-Train sentences took about 9 to 12 annotator-hours. For a similar cost, the rules therefore produce a better tagger, and unlike labels, they apply to any amount of unlabeled text. Finally, the errors AfriBERTa makes, such as confusing the verb \textit{no} (``to give'') with its auxiliary use (Table~\ref{tab:error_analysis}), are exactly the ambiguities that Tier~2 rules address.

\paragraph{Rules alone are insufficient}
The Rule-based Classifier reaches only 62.4\% accuracy (Macro F1: 0.584), although it uses the same linguistic knowledge as R2T. It is perfect on unambiguous, high-frequency words such as pronouns (F1 = 1.00), but it fails completely on context-dependent tags such as adjectives and determiners (F1 = 0.00). This gap shows that rules provide useful constraints but cannot resolve ambiguity on their own; a neural model is needed to learn the contextual patterns that do. R2T combines both.

\paragraph{BiLSTM vs.\ Transformer under rule-based training}
In the unsupervised setting, R2T-BiLSTM (0.968) outperforms R2T-Transformer (0.852) by a wide margin. One possible explanation is that the BiLSTM's sequential recurrence gives a stronger local signal, which fits our token-level loss. In contrast, the Transformer's global self-attention may dilute token-level rule constraints; for example, it systematically tags common verbs such as \textit{wani} (``to play'') as nouns. However, a simpler explanation is also possible: a 10-layer Transformer trained from scratch on 25,000 sentences may be too large for the amount of data. Our experiments do not separate these two explanations.

\paragraph{R2T as pre-training.}
After fine-tuning on only 50 labeled sentences, the Macro F1 of R2T-Transformer rises from 0.852 to 0.935, close to AfriBERTa trained on 300 (0.941). The effect is even larger for NER, where R2T-Transformer SFT-50 reaches 0.83 span F1 and outperforms AfriBERTa SFT-300 (0.79). These results suggest that rule-based pre-training gives the model a good starting point, so a small amount of labeled data is enough to fix the remaining errors. This two-stage approach (first learn from rules, then refine with labels) offers a practical way to build NLP tools in low-resource settings.

\section{Conclusion}
\label{sec:conclusion}

We presented Rule-to-Tag (R2T), a framework that builds a multi-tiered system of linguistic rules directly into the training objective of a neural sequence tagger. Our experiments on Zarma show two main strengths. First, for POS tagging, R2T-BiLSTM reaches 0.968 Macro F1 without labeled training data. It comes within 0.007 of a supervised BiLSTM-CRF and outperforms AfriBERTa fine-tuned on 300 labeled sentences, which answers RQ1. Second, for NER, R2T is a data-efficient pre-training method: after R2T pre-training, a model fine-tuned on 50 labeled sentences outperforms AfriBERTa fine-tuned on 300, which answers RQ2. Experiments on Bambara and French further suggest that the framework transfers to other languages. We release ZarmaPOS-Bench, ZarmaNER-600, and all trained models with this paper.

Beyond R2T itself, we return to the broader paradigm from Section~\ref{sec:introduction}, \textbf{principled learning (PrL)}: using explicit principles as a learning signal, rather than relying on example-based supervision alone. Instead of showing a model what the correct answer is, PrL provides a set of principles that encode the structure of the task; the model's goal is then to discover valid solutions that satisfy these principles. The principles can reach the model through different channels, such as a loss term or a reward. R2T uses the loss function and shows that this channel is viable: rules integrated as differentiable loss components can (to some extent) replace labeled data for dense tasks and provide a good initialization for sparse ones, all without auxiliary optimization and with no labeled data in the rule-based stage. We view R2T as a pilot implementation of PrL, and we believe this direction holds promise for other knowledge-intensive domains where expert principles are available but annotated examples are not.

\section{Limitations}
\label{sec:limitations}

We acknowledge several limitations that define the scope of this work.

First, our evaluation sets are small. The NER test set has only 100 sentences, and the two best POS models differ by less than their variation across seeds (0.975 vs.\ 0.968 Macro F1). These sizes reflect the cost of expert annotation in low-resource settings, but they limit how finely we can compare models.

Second, we do not ablate the individual loss terms, and we do not measure calibration. We therefore cannot say how much each term, including the OOV term, contributes to the results. We also applied supervised fine-tuning only to R2T-Transformer, so we do not know how much R2T-BiLSTM would gain from it.

Third, R2T introduces four loss-weight hyperparameters ($\alpha, \beta, \gamma, \delta$), which we set on Rule-Dev. Individual training runs are cheap compared to pre-training large language models, but searching for good weights still needs compute that may strain small budgets.

Fourth, R2T depends on human-crafted rules and on a small labeled set to refine them. Rule writing took about 4 hours for Zarma (one trained native speaker and one NLP researcher) and about 2.75 hours for Bambara, where we built on existing Daba morphemic rules. By contrast, annotating the 300 Gold-Train sentences took about 9 to 12 annotator-hours, excluding adjudication. The human effort is therefore similar, but R2T presupposes access to linguistic expertise, and rule writing may take longer for morphologically complex languages.

Fifth, our cross-lingual comparisons are limited. For Bambara, we compare against a released tagger rather than a baseline trained on our data, and for French we report R2T only.

Sixth, we deliberately exclude large language models (beyond embedding-based models) from our comparisons. These models do not fit our target setting: we focus on accessible, reproducible methods that researchers and communities can train and deploy on consumer-grade hardware.

Finally, we test R2T on three languages: Zarma (Songhay), Bambara (Mande), and French (Romance). Its performance on languages with rich morphology, such as agglutinative or polysynthetic languages, remains an open question, and such languages may need richer morphological or syntactic tiers. More broadly, R2T tests only one PrL channel (the loss) on one task family (sequence tagging), so other channels and tasks remain future work.

\bibliography{custom}

\appendix

\section{Technical Details}
\label{sec:technical_details}

This section gives the architectural details and training hyperparameters needed for full reproducibility.

\subsection{Model Architectures}
\label{ssec:appendix_arch}

Both R2T models share the same input representation (concatenated FastText and character embeddings) and the same rule-informed loss function. They differ only in their core sequence encoder.

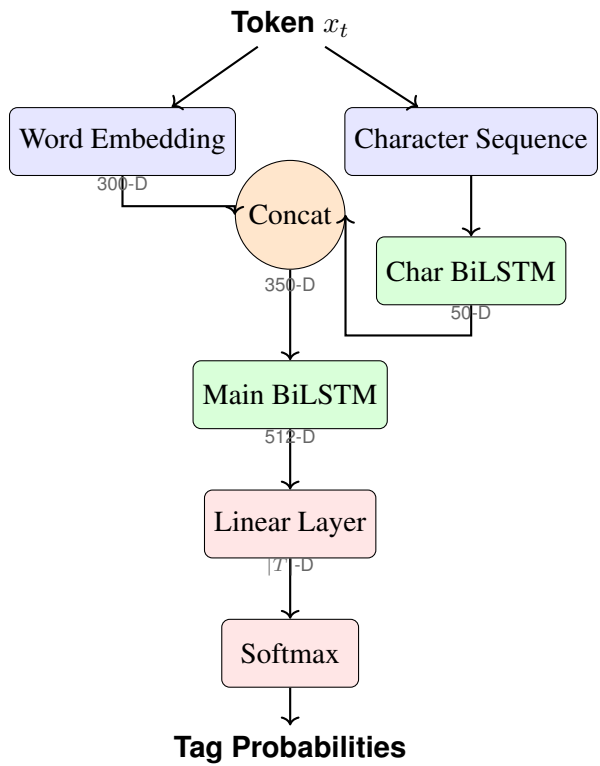
\begin{figure}[h]
    \centering
    \begin{tikzpicture}[
        node distance=1.5cm and 1cm,
        emb/.style={rectangle, draw, fill=blue!10, minimum height=0.9cm, minimum width=1.8cm, rounded corners=3pt},
        lstm/.style={rectangle, draw, fill=green!15, minimum height=0.9cm, minimum width=2.2cm, rounded corners=3pt},
        op/.style={circle, draw, fill=orange!20, minimum size=0.8cm},
        token/.style={rectangle, draw=none, font=\sffamily\bfseries},
        dim/.style={rectangle, draw=none, font=\sffamily\scriptsize, text=gray!80!black}
    ]
    \node[token] (word) at (0,0) {Token $x_t$};
    \node[emb] (word_emb) [below left=0.8cm and -0.2cm of word] {Word Embedding};
    \node[emb] (char_seq) [below right=0.8cm and -0.2cm of word] {Character Sequence};
    \node[lstm] (char_lstm) [below=0.8cm of char_seq] {Char BiLSTM};
    \node[op] (concat) [below=1.5cm of word] {Concat};
    \node[lstm] (main_lstm) [below=1.2cm of concat] {Main BiLSTM};
    \node[emb] (linear) [below=0.8cm of main_lstm, fill=red!10] {Linear Layer};
    \node[emb] (softmax) [below=0.8cm of linear, fill=red!10] {Softmax};
    \node[token] (tag) [below=0.5cm of softmax] {Tag Probabilities};

    \draw[->, thick] (word) -- (word_emb);
    \draw[->, thick] (word) -- (char_seq);
    \draw[->, thick] (char_seq) -- (char_lstm);
    \draw[->, thick] (word_emb.south) -- ++(0,-0.4) -| (concat.west);
    \draw[->, thick] (char_lstm.south) -- ++(0,-0.4) -| (concat.east);
    \draw[->, thick] (concat) -- (main_lstm);
    \draw[->, thick] (main_lstm) -- (linear);
    \draw[->, thick] (linear) -- (softmax);
    \draw[->, thick] (softmax) -- (tag);
    
    \node[dim] at ($(word_emb.south) + (0,-0.1)$) {300-D};
    \node[dim] at ($(char_lstm.south) + (0,-0.1)$) {50-D};
    \node[dim] at ($(concat.south) + (0,-0.2)$) {350-D};
    \node[dim] at ($(main_lstm.south) + (0,-0.1)$) {512-D};
    \node[dim] at ($(linear.south) + (0,-0.1)$) {$|T|$-D};
    \end{tikzpicture}
    \caption{Architecture of the R2T-BiLSTM model.}
    \label{fig:bilstm_arch}
\end{figure}

\textbf{R2T-BiLSTM.} Our recurrent model (Figure~\ref{fig:bilstm_arch}) takes a 350-dimensional input vector per token: a 300-dimensional FastText word embedding concatenated with a 50-dimensional character-level embedding. The character embedding comes from a single-layer character BiLSTM with 25 hidden units per direction. This 350-dimensional vector feeds into the main token-level BiLSTM (one layer, 256 hidden units per direction), which produces a 512-dimensional context-aware representation. A linear layer then maps this representation to tag logits.

\begin{figure}[h]
    \centering
    \scriptsize
    \begin{tikzpicture}[
        node distance=1.2cm and 0.8cm,
        >=stealth,
        emb/.style={rectangle, draw, fill=blue!10, minimum height=0.8cm, minimum width=1.6cm, rounded corners=2pt},
        bilstm/.style={rectangle, draw, fill=green!15, minimum height=0.8cm, minimum width=2.2cm, rounded corners=2pt},
        trans/.style={rectangle, draw, fill=purple!15, minimum height=0.8cm, minimum width=2.4cm, rounded corners=2pt},
        op/.style={circle, draw, fill=orange!20, minimum size=0.7cm},
        token/.style={rectangle, draw=none, font=\sffamily\bfseries},
        dim/.style={rectangle, draw=none, font=\sffamily\scriptsize, text=gray!80!black}
    ]
    \node[token] (word) at (0,0) {Token $x_t$};
    \node[emb] (word_emb) [below left=0.6cm and -0.1cm of word] {Word Embedding};
    \node[emb] (char_seq) [below right=0.6cm and -0.1cm of word] {Character Sequence};
    \node[bilstm] (char_lstm) [below=0.6cm of char_seq] {Char BiLSTM};
    \node[op] (concat) [below=1.2cm of word] {Concat};
    
    \node[emb, fill=red!10] (projection) [below=1.0cm of concat] {Input Projection};
    \node[op] (add) [below=0.8cm of projection] {+};
    \node[emb, fill=yellow!20] (pos_enc) [right=0.8cm of add] {Positional Encoding};
    \node[trans] (transformer) [below=0.8cm of add] {Transformer Encoder};
    \node[emb, fill=red!10] (linear) [below=0.6cm of transformer] {Linear Layer};
    \node[emb, fill=red!10] (softmax) [below=0.6cm of linear] {Softmax};
    \node[token] (tag) [below=0.4cm of softmax] {Tag Probabilities};

    \draw[->, thick] (word) -- (word_emb);
    \draw[->, thick] (word) -- (char_seq);
    \draw[->, thick] (char_seq) -- (char_lstm);
    \draw[->, thick] (word_emb.south) -- ++(0,-0.3) -| (concat.west);
    \draw[->, thick] (char_lstm.south) -- ++(0,-0.3) -| (concat.east);
    \draw[->, thick] (concat) -- (projection);
    \draw[->, thick] (projection) -- (add);
    \draw[->, thick] (pos_enc) -| (add);
    \draw[->, thick] (add) -- (transformer);
    \draw[->, thick] (transformer) -- (linear);
    \draw[->, thick] (linear) -- (softmax);
    \draw[->, thick] (softmax) -- (tag);
    
    \node[dim] at ($(concat.south) + (0,-0.15)$) {350-D};
    \node[dim] at ($(projection.south) + (0,-0.1)$) {768-D};
    \node[dim] at ($(transformer.south) + (0,-0.1)$) {768-D};
    \node[dim] at ($(linear.south) + (0,-0.1)$) {$|T|$-D};
    \end{tikzpicture}
    \caption{Architecture of the R2T-Transformer model.}
    \label{fig:transformer_arch}
\end{figure}
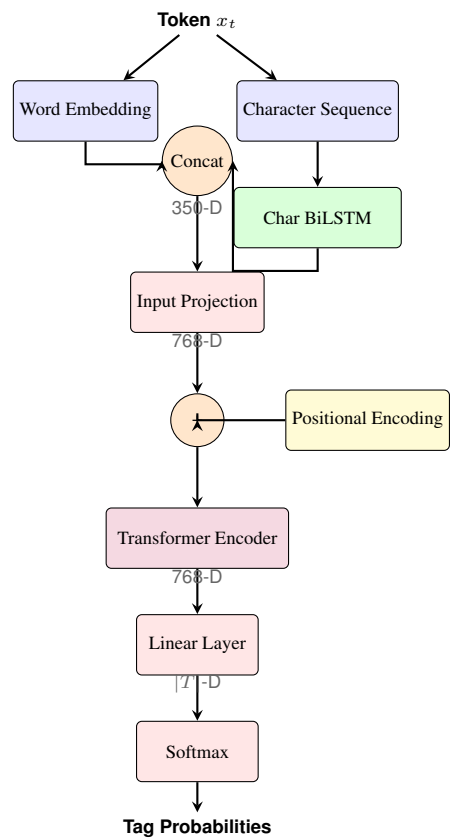

\textbf{R2T-Transformer.} Our attention-based model (Figure~\ref{fig:transformer_arch}) replaces the recurrent core with a Transformer encoder. A linear layer projects the 350-dimensional input vector to 768 dimensions, and we add sinusoidal positional encodings. A 10-layer Transformer encoder (6 self-attention heads, 3072 feed-forward units per layer) processes this representation, and a linear layer maps its 768-dimensional output to tag logits.

\textbf{NER models.} For the NER experiments (Section~\ref{ssec:ner_results}), we use the same architectures and change only the final linear layer, so that it outputs the BIO-format NER tagset instead of the POS tagset.

\subsection{Training Hyperparameters}
\label{ssec:appendix_hyper}

Table~\ref{tab:hyperparams} lists the hyperparameters for all models.

\begin{table*}[ht]
\centering
\small
\begin{tabular}{l|ccccc}
\hline
\textbf{Hyperparameter} & \textbf{R2T-BiLSTM} & \textbf{R2T-Transformer} & \textbf{AfriBERTa} & \textbf{XLM-RoBERTa} & \textbf{BiLSTM-CRF} \\
\hline \hline
\multicolumn{6}{c}{\textit{Model Architecture}} \\
\hline
Word Embedding Dim & 300 (FastText) & 300 (FastText) & 768 & 768 & 100 (Learned) \\
Char Embedding Dim & 50 & 50 & N/A & N/A & 25 \\
Hidden Dim & 256 (x2) & 768 & 768 & 768 & 128 (x2) \\
Num.\ Layers & 1 & 10 & 12 & 12 & 1 \\
Num.\ Heads & N/A & 6 & 12 & 12 & N/A \\
Feed-Forward Dim & N/A & 3072 & 3072 & 3072 & N/A \\
Dropout & 0.3 & 0.1 & 0.1 & 0.1 & 0.5 \\
\hline
\multicolumn{6}{c}{\textit{Training \& Fine-Tuning}} \\
\hline
Optimizer & Adam & Adam & AdamW & AdamW & Adam \\
Learning Rate & 1e-3 & 5e-5 & 2e-5 & 2e-5 & 1e-3 \\
Batch Size & 256 & 64 & 16 & 16 & 16 \\
Epochs & 30 & 30 (unsup.) / 20 (SFT) & 10 & 10 & 50 \\
Weight Decay & 1e-5 & 1e-5 & 0.01 & 0.01 & 1e-4 \\
Max Grad Norm & 1.0 & 1.0 & 1.0 & 1.0 & 1.0 \\
\hline
\multicolumn{6}{c}{\textit{R2T Loss Weights}} \\
\hline
$\alpha$ (Lexical) & 0.85 & 0.85 & N/A & N/A & N/A \\
$\beta$ (Syntactic) & 0.08 & 0.08 & N/A & N/A & N/A \\
$\gamma$ (Distributional) & 0.02 & 0.02 & N/A & N/A & N/A \\
$\delta$ (OOV) & 0.05 & 0.05 & N/A & N/A & N/A \\
\hline
\end{tabular}
\caption{Training and architectural hyperparameters for all models.}
\label{tab:hyperparams}
\end{table*}

\section{ZarmaPOS-Bench}
\label{sec:zarmapos_bench}

Low-resource NLP research lacks large annotated datasets for tasks such as POS tagging~\citep{Khurana_2022}. To help close this gap and support further research on Zarma, we introduce \textbf{ZarmaPOS-Bench}, the first large-scale POS-tagged corpus for Zarma. Our own model produced its tags, so it is a silver-standard corpus: researchers should evaluate new taggers on the gold data (Section~\ref{ssec:bench_gold}), not on the silver corpus.

\subsection{Motivation}
\label{ssec:bench_motivation}

A large gold-standard corpus is ideal but often infeasible in low-resource settings. A practical alternative is a high-quality silver-standard corpus produced by automatic annotation. Since R2T-BiLSTM reaches 98.2\% accuracy without labeled training data, it is a good candidate for this purpose. ZarmaPOS-Bench gives the research community a large resource that, while imperfect, is good enough to support new research and applications for Zarma.

\subsection{Data Curation and Annotation}
\label{ssec:bench_curation}

We built ZarmaPOS-Bench from the Feriji dataset~\citep{keita-etal-2024-feriji}. We processed 46,064 rows by splitting multi-sentence entries and tokenizing with \texttt{wordpunct\_tokenize}. Our R2T-BiLSTM model then tagged each sentence, which produced a silver-standard corpus of about 55,000 sentences and 1,005,295 tokens in JSONL format.

\subsection{Dataset Statistics}
\label{ssec:bench_stats}

ZarmaPOS-Bench contains about \textbf{55,000 sentences} and more than \textbf{1,000,000 tokens}. Each JSONL line contains three fields:
\begin{itemize}
    \item \texttt{text}: The original, untokenized sentence.
    \item \texttt{tokens}: A list of tokenized strings.
    \item \texttt{tags}: A parallel list of predicted POS tags.
\end{itemize}

An example entry:
\begin{verbatim}
{
  "text": "Waybora di alboro.",
  "tokens": ["Waybora", "di",
             "alboro", "."],
  "tags": ["NOUN", "VERB",
           "NOUN", "PUNCT"]
}
\end{verbatim}

Table~\ref{tab:tag_distribution} shows the tag distribution. Nouns, pronouns, auxiliaries, and verbs dominate, which matches typical linguistic patterns.

\begin{table}[h]
\centering
\tiny
\begin{tabular}{lrr}
\hline
\textbf{POS Tag} & \textbf{Count} & \textbf{Frequency (\%)} \\
\hline
NOUN & 241,274 & 24.0 \\ 
PRON & 168,153 & 16.7 \\ 
AUX & 162,423 & 16.2 \\ 
PUNCT & 156,019 & 15.5 \\ 
VERB & 146,118 & 14.5 \\ 
PART & 81,387 & 8.1 \\ 
ADJ & 26,777 & 2.7 \\ 
DET & 21,340 & 2.1 \\ 
OTHER & 1,804 & 0.2 \\ 
\hline
Total & 1,005,295 & 100.0 \\ 
\hline
\end{tabular}
\caption{Tag distribution in ZarmaPOS-Bench. The \textsc{other} tag marks very low-confidence tokens.}
\label{tab:tag_distribution}
\end{table}

\subsection{Gold Standard Data}
\label{ssec:bench_gold}

ZarmaPOS-Bench is silver-standard and inevitably contains errors, probably concentrated on subtle adjective--noun ambiguities and rare out-of-domain words. We therefore release it with the gold splits described in Section~\ref{ssec:data}. The gold data serves three purposes: (1)~Gold-Test is the benchmark for evaluating new Zarma POS taggers; (2)~Gold-Train is a supervised fine-tuning set that corrects the silver model's systematic errors (as in our SFT-50 experiments); and (3)~the gold data can seed active learning pipelines.

The full dataset is available at: \url{https://huggingface.co/datasets/27Group/Zarma_POS}.

\section{Generalization to Bambara}
\label{sec:bambara_experiment}

We test R2T on Bambara, a Manding language of the Mande family spoken in West Africa. Bambara poses different grammatical challenges than Zarma, including a greater role for tone and more complex verb-auxiliary constructions. Table~\ref{tab:cross_lingual} gives the main results; this appendix gives the full details.

\subsection{Setup}
\label{ssec:bambara_setup}

\textbf{Rules.} We wrote a multi-tiered rule system for Bambara based on Daba morphemic rules~\citep{daba}: about 100 unambiguous lexical entries, rules for ambiguous function words (\textit{ye}, \textit{ka}, \textit{ma}), morphological suffixes (such as the plural \textit{-w}), and core syntactic constraints. Writing the rules took about 2.75 hours.

\textbf{Data.} We use 864 unlabeled Bambara sentences from the SMOL dataset~\citep{caswell2025smol} for unsupervised training and 1,000 sentences for evaluation.

\textbf{Model.} We use a hybrid architecture that combines a pre-trained T5 encoder (\textbf{t5-small})~\citep{2020t5} with our BiLSTM tagger head. The T5 encoder produces contextual embeddings, which feed into the BiLSTM. We train the whole model with only the Bambara rule system and the unlabeled corpus.

\textbf{Baseline.} We compare against the released Masakhane AfroXLMR tagger,\footnote{\url{https://huggingface.co/masakhane/bambara-pos-tagger-afroxlmr}} a supervised model fine-tuned on manually annotated Bambara data. We use the released model as is.

\subsection{Analysis}
\label{ssec:bambara_analysis}

R2T outperforms the supervised Masakhane tagger by 0.13 Macro F1 (0.91 vs.\ 0.78) and by 10.2 points in accuracy (92.7\% vs.\ 82.5\%). Table~\ref{tab:bambara_error_analysis} analyzes the baseline's errors. These errors are systematic: the baseline gives the same tag to both instances of the ambiguous function word \textit{ye}, confuses adjectives with nouns, and tags core vocabulary inconsistently. They are exactly the error types that R2T's tiered rules target.

\begin{table*}[h]
\centering
\scriptsize 
\begin{tabular}{p{3.5cm}|p{4.5cm}|p{7cm}}
\hline
\textbf{Error Category} & \textbf{Example Sentence \& Prediction} & \textbf{Analysis \& R2T Advantage} \\
\hline \hline

\textbf{Ambiguity of function words} & \textit{I ye wulu ye.} (You saw a dog.) \newline \textbf{Pred:} \textit{ye} $\rightarrow$ \textsc{part}, \textit{ye} $\rightarrow$ \textsc{part} \newline \textbf{Gold:} \textit{ye} $\rightarrow$ \textsc{aux}, \textit{ye} $\rightarrow$ \textsc{verb} & The baseline assigns the same tag to both instances. R2T's Tier~2 rule \texttt{\{ye: [AUX, VERB, PART]\}} combined with syntactic constraints enables positional disambiguation. \\
\hline

\textbf{Adjective--noun confusion} & \textit{C\textepsilon{} surun b\textepsilon{} taa.} (The short man is going.) \newline \textbf{Pred:} \textit{surun} $\rightarrow$ \textsc{noun} \newline \textbf{Gold:} \textit{surun} $\rightarrow$ \textsc{adj} & The baseline fails to learn the \textsc{noun + adj} pattern. R2T's syntactic rule \texttt{(NOUN, ADJ): 1.0} encourages correct predictions and generalizes the pattern. \\
\hline

\textbf{Inconsistent core vocabulary} & \textit{Ji b\textepsilon{} min.} (Water is being drunk.) \newline \textbf{Pred:} \textit{min} $\rightarrow$ \textsc{pron} \newline \textbf{Gold:} \textit{min} $\rightarrow$ \textsc{verb} & The baseline misclassifies a common verb. R2T's Tier~1 lexicon defines ``min'' as \textsc{verb}, making this error unlikely. \\
\hline

\end{tabular}
\caption{Qualitative error analysis of the Masakhane baseline on Bambara.}
\label{tab:bambara_error_analysis}
\end{table*}

\section{Generalization to French}
\label{ssec:french_experiment}

To test whether R2T transfers beyond African language families, we run a stress test on French under simulated low-resource conditions.

\textbf{Data} We sample 1,000 unlabeled sentences from the Universal Dependencies French-GSD dataset for unsupervised training and evaluate on the standard UD French test set. We use only R2T-BiLSTM, since all other baselines are already heavily pre-trained on French.

\textbf{Rules} We wrote 15 basic rules in under one hour: lexical rules for common function words (\textit{le, la, les} as determiners; \textit{et, ou} as conjunctions) and morphological rules (for example, words ending in \textit{-ment} are likely adverbs).

\textbf{Results} R2T-BiLSTM reaches 0.941 Macro F1 and 95.8\% word-level accuracy (Table~\ref{tab:cross_lingual}). French has abundant annotated data, so R2T is not needed for it in practice. Still, this result suggests that the framework is not specific to Zarma or Bambara and can transfer to other languages when basic rules can be written and data is limited.

\section{Rule Creation Details}
\label{sec:rule_creation}

For both languages, we refined the rules step by step based on errors on the Rule-Dev set. For Zarma, the initial rules led the model to tag certain verbs as nouns (for example, \textit{wani}), so we added specific lexical entries to Tier~1. For Bambara, tone-related ambiguities (for example, \textit{ye} as \textsc{aux} or \textsc{verb}) required a larger Tier~2 lexicon. Each iteration consisted of training an R2T model, analyzing its errors on Rule-Dev, and updating the rules. This cycle typically took 2--3 iterations before we froze the rule set.

\section{Additional Figures}
\label{sec:additional_figures}

\subsection{Confusion Matrix for Zarma POS Tagging}

\begin{figure}[h]
  \centering
  \includegraphics[width=\columnwidth]{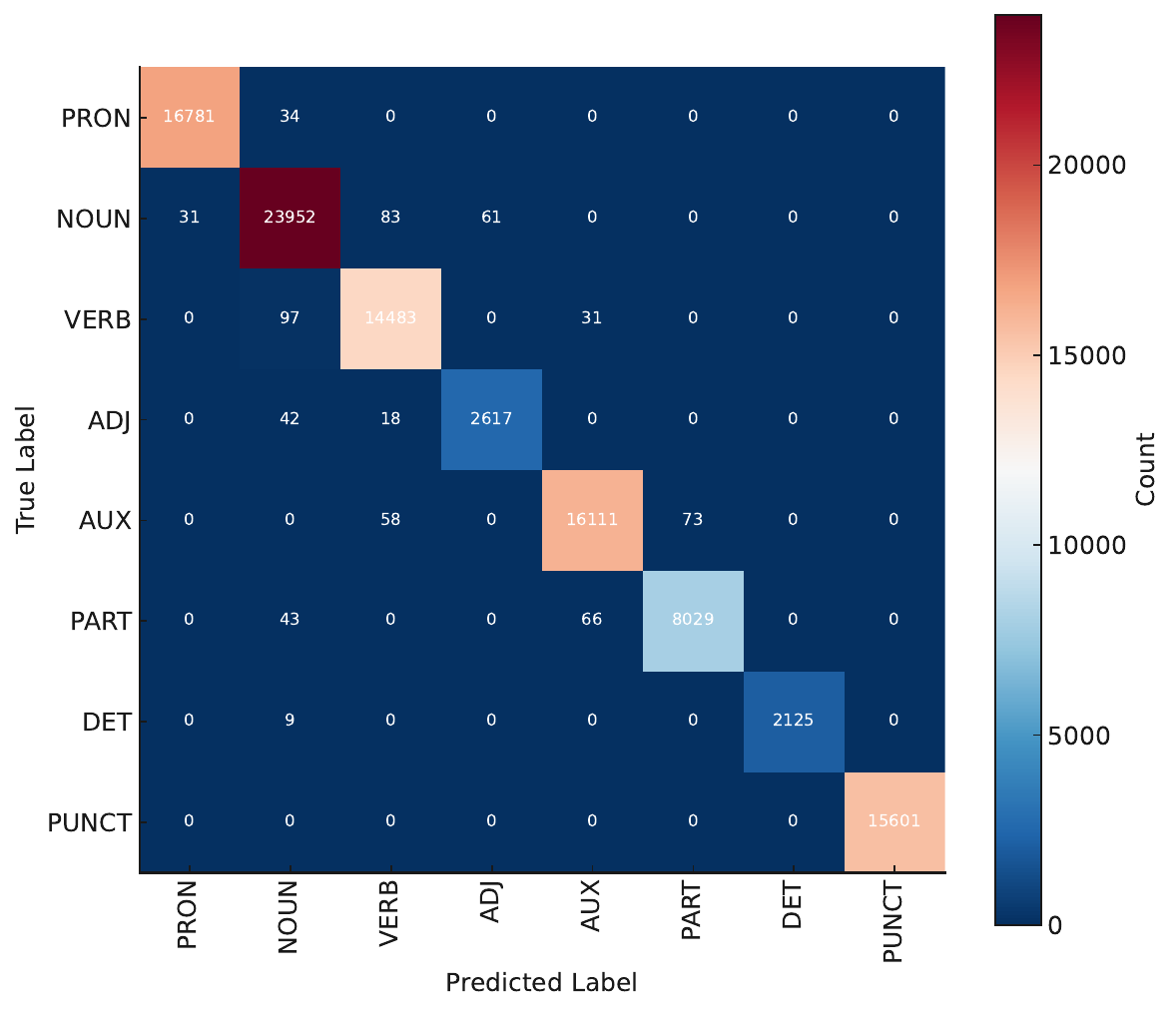}
  \caption{Confusion matrix for the R2T-BiLSTM POS tagger on the 1,000-sentence Zarma Gold-Test set.}
  \label{fig:confusion_matrix}
\end{figure}

Figure~\ref{fig:confusion_matrix} visualizes the predictions of R2T-BiLSTM. The strong diagonal indicates high accuracy across all tags. The few off-diagonal entries show minor confusions between \textsc{noun} and \textsc{verb} and between \textsc{part} and \textsc{aux}. Both are grammatically motivated ambiguities rather than random errors, and this pattern is consistent with the qualitative analysis in Table~\ref{tab:error_analysis}.

\subsection{Full Error Analysis for Zarma POS}

Table~\ref{tab:full_error_analysis} extends the representative errors in Table~\ref{tab:error_analysis} with more examples from the R2T-Transformer variants.

\begin{table*}[ht]
\centering
\scriptsize
\begin{tabular}{p{2.5cm}|p{2.5cm}|p{4cm}|p{5.5cm}}
\hline
\textbf{Model} & \textbf{Error Type} & \textbf{Example} & \textbf{Analysis} \\
\hline \hline

R2T-Transformer & Verb misclassification & \textit{Iri ga wani.} \newline Pred: \textit{wani} $\rightarrow$ \textsc{noun} \newline Gold: \textit{wani} $\rightarrow$ \textsc{verb} & The Transformer's global attention dilutes the token-level lexical rule signal for ``wani'' (to play). \\
\hline

R2T-Transformer & Disambiguation failure & \textit{Ay no a se moo.} \newline Pred: \textit{no} $\rightarrow$ \textsc{aux} \newline Gold: \textit{no} $\rightarrow$ \textsc{verb} & Same ambiguity as AfriBERTa: the Tier~2 rule alone does not suffice for the Transformer without supervised examples. \\
\hline

R2T-Transformer SFT-50 & Residual ambiguity & \textit{Ay no a se moo.} \newline Pred: \textit{no} $\rightarrow$ \textsc{aux} \newline Gold: \textit{no} $\rightarrow$ \textsc{verb} & SFT corrected most errors, but the 50 fine-tuning sentences lacked diverse examples of ``no'' as a main verb. \\
\hline

R2T-Transformer SFT-50 & Residual adj--noun confusion & \textit{Wayboro hanno na ay no gaasi.} \newline Pred: \textit{hanno} $\rightarrow$ \textsc{noun} \newline Gold: \textit{hanno} $\rightarrow$ \textsc{adj} & The fine-tuning set likely lacked examples of ``hanno'' (beautiful) as an adjective to override the pre-existing bias. \\
\hline

\end{tabular}
\caption{Extended error analysis for R2T-Transformer variants on Zarma POS tagging.}
\label{tab:full_error_analysis}
\end{table*}

\end{document}